\documentclass[conference]{IEEEtran}

\IEEEoverridecommandlockouts

\usepackage[numbers,sort&compress]{natbib}
\usepackage{url}

\usepackage{graphicx}
\usepackage{color,hyperref}
\definecolor{blue}{rgb}{0.0,0.0,1.0}
\usepackage[vlined, ruled, boxed]{algorithm2e}
\usepackage{subcaption}

\usepackage{multirow}
\usepackage{diagbox}

\usepackage{amssymb,amsmath,amsthm}
\usepackage{amsfonts}
\usepackage{array}
\usepackage[pagewise]{lineno}
\usepackage{eurosym}
\usepackage{bbm}
\usepackage{bm}

\usepackage{algorithm2e}

\usepackage[utf8]{inputenc}
\usepackage[T1]{fontenc}

\usepackage[colorinlistoftodos]{todonotes}

\makeatletter
\newcommand{\linebreakand}{%
  \end{@IEEEauthorhalign}
  \hfill\mbox{}\par
  \mbox{}\hfill\begin{@IEEEauthorhalign}
}
\makeatother

\usepackage{todonotes}
\begin{document}

% Title page.
\title{A domain adaptation neural network for digital twin-supported fault diagnosis} 

\author{\IEEEauthorblockN{Zhenling Chen}
\IEEEauthorblockA{CentraleSup\'{e}lec, \\ Universit\'{e} Paris-Saclay, \\ Gif-sur-Yvette, 91190,~France}
\and
\IEEEauthorblockN{Haiwei Fu}
\IEEEauthorblockA{CentraleSup\'{e}lec, \\ Universit\'{e} Paris-Saclay, \\ Gif-sur-Yvette, 91190,~France}
% \makeatother
% \IEEEauthorblockA{Arts et Matiers Institute of Technology \\ LAMPA \\ F-53810 Change, France}

\linebreakand 
%\IEEEauthorblockN{Lama Itani}
%\IEEEauthorblockA{Mathworks France, \\ Meudon, France}
%\and
\IEEEauthorblockN{Zhiguo Zeng}
\IEEEauthorblockA{Chair on Risk and Resilience of Complex Systems, \\ Laboratoie Genie Industriel, CentraleSup\'{e}lec, \\ Universit\'{e} Paris-Saclay, 91190, Gif-sur-Yvette,~France}
}

\maketitle

\begin{abstract}
Digital twins offer a promising solution to the lack of sufficient labeled data in deep learning-based fault diagnosis by generating simulated data for model training. However, discrepancies between simulation and real-world systems can lead to a significant drop in performance when models are applied in real scenarios. To address this issue, we propose a fault diagnosis framework based on Domain-Adversarial Neural Networks (DANN), which enables knowledge transfer from simulated (source domain) to real-world (target domain) data.

We evaluate the proposed framework using a publicly available robotics fault diagnosis dataset, which includes 3,600 sequences generated by a digital twin model and 90 real sequences collected from physical systems. The DANN method is compared with commonly used lightweight deep learning models such as CNN, TCN, Transformer, and LSTM. Experimental results show that incorporating domain adaptation significantly improves the diagnostic performance. For example, applying DANN to a baseline CNN model improves its accuracy from 70.00\% to 80.22\% on real-world test data, demonstrating the effectiveness of domain adaptation in bridging the sim-to-real gap.
\footnote{Code and datasets available at: \href{https://github.com/JialingRichard/Digital-Twin-Fault-Diagnosis}{https://github.com/JialingRichard/Digital-Twin-Fault-Diagnosis}}
\end{abstract}

\begin{IEEEkeywords}
predictive maintenance, fault diagnosis, digital failure twin, domain adaptation neural network (DANN)
\end{IEEEkeywords}


% Glossaries
% Print the glossary
%\setlength{\glsdescwidth}{.8\textwidth}
%\printglossary[type=\acronymtype,style=long]
%\printnomenclature[1.5cm]

% \IEEEpeerreviewmaketitle
%\thispagestyle{empty} % no page number for the first page

%\linenumbers

\section{Introduction}

Fault diagnosis aims at identifying the cause of a failure from observational data from sensors \cite{zhang2023digital}. One of the major challenge in fault diagnosis is that the state-of-the-art deep learning-based models often require large amount of data. It is, however, often difficult to obtain these data in practice \cite{zhong2023overview}. Digital twin technology combines physical entity with its digital representation. It can accurately reproduce the scenes in the physical world in the virtual environment, providing great convenience for the analysis, optimization and control of physical system \cite{juarez2021digital}. Using digital twins to generate simulated failure data and train a deep learning model for fault diagnosis has become a promising approach to solve the data insufficiency issue of fault diagnosis. 

There are already some existing works in applying digital twins for fault diagnosis. For example, Jain et al. \cite{jain2019digital} proposed a digital twin-based fault diagnosis framework that utilizes the digital twin model to simulate system behavior and identify fault patterns in distributed photovoltaic systems, Wang et al. \cite{wang2019digital} proposed a digital twin-based fault diagnosis framework that integrates sensor data and physical models to detect and diagnose faults in rotating machinery within smart manufacturing systems. Yang et al.\cite{yang2023digital} proposed a digital twin-driven fault diagnosis method that combines virtual and real data to diagnose composite faults, where the digital twin generates virtual samples to compensate for the scarcity of fault samples in real systems. Most of these existing works assume that condition-monitoring data are availalbe on the same level as the component being diagnosed. In practice, however, deploying sensors at the component level is often difficult. One has to rely on system-level condition-monitoring data to infer the component-level failure modes \cite{ran2019survey}. In one of our previous works \cite{court2024use}, we developed a digital twin model of a robot and use it to generate simulated failure data for fault diagnosis. Testing data are collected from a real robot with different injected failures to test the performance of the developed model. 

The existing works share a common assumption: The digital twin model can accurately predict the actual behavior of the component under test. However, in practice, the digital twin model is not always accurate. Then, the fault diagnosis model trained on simulation data often suffers from poor performance when applied to real data, due to the imprecision of the simulation model. To address this issue, we propose a Domain Adversarial Neural Network (DANN)-based framework for digital twin-supported fault diagnosis. Through the DANN \cite{ganin2015unsupervised}, the developed model is able to learn useful features from the simulated data even the simulation does not exactly match the reality. We also performed a benchmark study by comparing the performance of the developed model with other state-of-the-art deep learning models, including LSTM \cite{hochreiter1997long}, Transformer \cite{vaswani2017attention}, CNN \cite{lecun1989handwritten} and TCN \cite{bai2018empirical}. The main contributions of this paper are:
\begin{itemize}
\item We propose a novel DANN-based framework for digital twin-supported fault diagnosis. 

\item We present an open-source dataset for digital twin-supported fault diagnosis. The dataset include simulated training data and real test data.

\item We conducted a detailed benchmark study where the performance of the developed model is compared with four other state-of-the-art deep learning models.
\end{itemize}

\section{Digital twin model and dataset description}

\begin{figure}[ht]
  \centering
  \includegraphics[width=.46\textwidth]{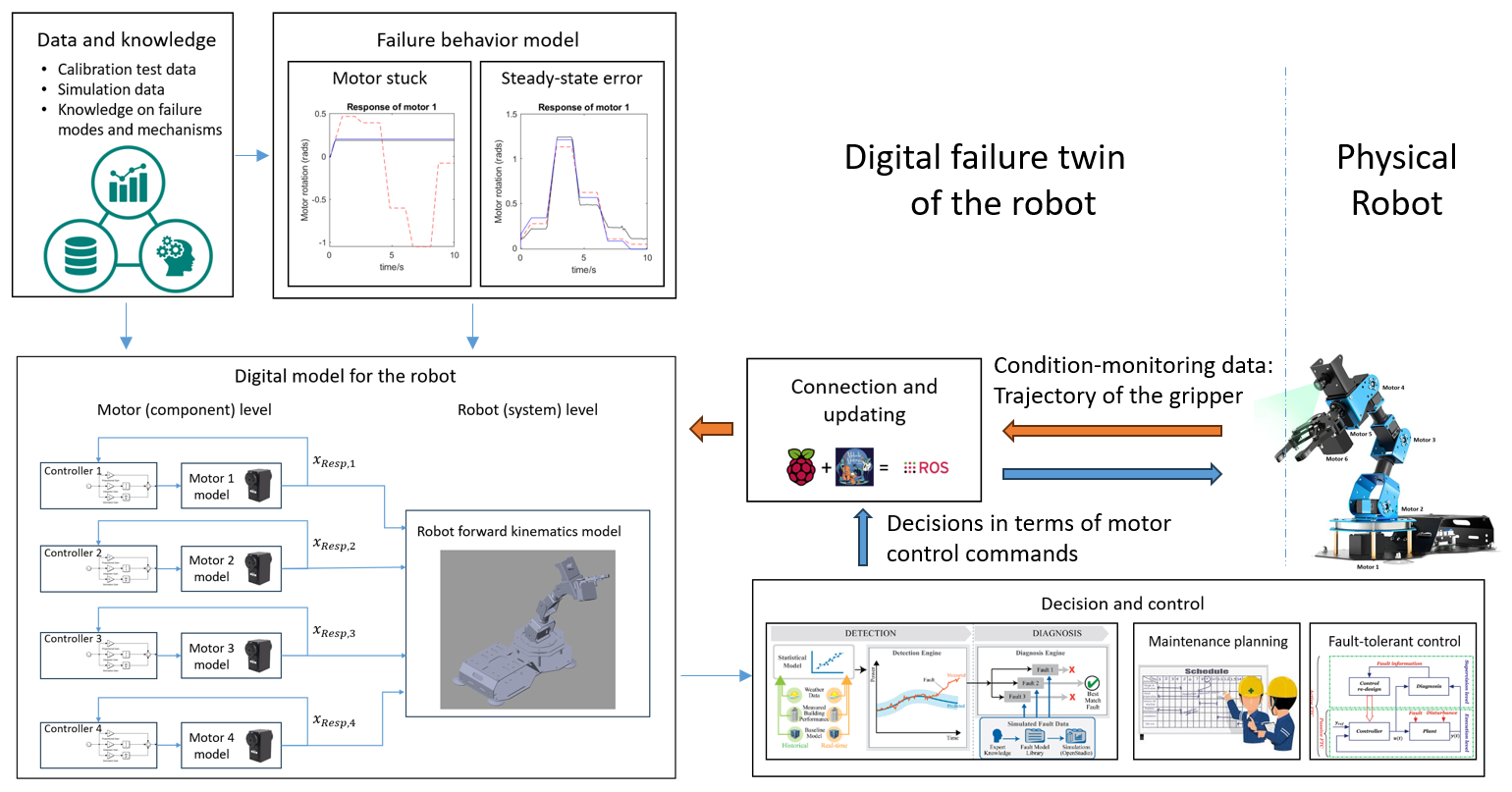}
  \caption{The fault diagnosis in digital twin for robot \cite{court2024use}.}
  \label{fig:dtrRobot}
\end{figure}

In this paper, we consider the open source dataset for digital twin-supported fault diagnosis we developed previously in \cite{court2024use}. The dataset is created based on the digital failure twin model of a robot, as shown in Fig. \ref{fig:dtrRobot}. 

A digital twin model is a simulation model used to simulate the failure behavior of the robot and connect to the physical entity to reflect its real-time states. The robot comprises of six motors. We monitor the trajectory of the end-effector and the control commands of each motor. The goal of the fault diagnosis is to use the condition-monitoring data to infer the failure modes of the four out of the six motors. Each motor might subject to two failure modes, i.e., stuck and steady-state error.

The digital failure twin model is built as a two-layer model. On the motor level, we model the dynamics of the motor and its controller. Then, the response of each motor is fed into a forward kinematics model, which allows simulating the end-effector trajectory from the postions of the motors. The stuck and steady-state error can be simulated by changing the response of each motor, as shown in Fig. \ref{fig:dtrRobot}.

To generate the training dataset, we generate $400$ random trajectories, and simulate the $9$ classes (one normal state and eight failure states where each motor could be in either one of the two failure modes) under each trajectory. Each sample contains records spanning 1000 time steps. Then, we collect test data by randomly simulate 90 trajectories following the same protocals. 

In the original work \cite{court2024use}, an LSTM was trained on the simulation dataset and applied to dignose the failures on the real robot. The results showed that, although the trained model performed well on the validation set (seperated from training data, but still from simulation), it performs poorly on the real testing dataset ($96\%$ V.S. $69\%$). The main reason is that the simulation model does not match exactly the behavior of the real robot. In this paper, we intend to address this issue through transfer learning.

\section{Existing transfer learning models}

Prevalent deep learning-based models show great success in both academia and industry \cite{he2017deep}. For example, Convolutional Neural Networks (CNN) use in automated fault detection for machinery vibrations \cite{xia2017fault}, Recurrent Neural Networks (RNN) for example, LSTM have proven useful in diagnosing faults based on time series data \cite{shi2022planetary}. In current work, Plakias et al. combined the dense convolutional blocks and the attention mechanism to develop a new attentive dense CNN for fault diagnosis \cite{plakias2020fault}. Although these methods
can achieve high performance in fault diagnosis, the application of these methods is usually under the assumption that test data and train data come from the same data distribution. Also, the current deep learning-based models are under the Independent Identically Distribution (i.i.d.). 

As we discussed before, the data generated from a digital twin might not exactly match the actual behavior in the physical entity. As a result, the distribution of training and testing dataset cannot be assumed as i.d.d., due to steady-state errors that cause in friction or other mechanical effects and real time faults that can big impact the results. In this paper, we use transfer learning methods to deal with source domain and target domain alignment in digital twin in data distribution. 

To solve the issue of data distribution discrepancy, various domain adaptation techniques in transfer learning have been introduced for diagnosing bearing faults \cite{li2022perspective, zhiyi2020transfer, cao2022unsupervised}.  Transfer learning can also be used to learn knowledge from  source domain for fault diagnosis on a different target domain. Applications of transfer learning in fault diagnosis include representation
adaptation \cite{guo2018deep, pang2019cross, xiao2019domain, yang2019intelligent}, parameter transfer \cite{he2019improved, kim2019new, shao2018highly}, adversarial-based domain
adaptation \cite{cheng2020wasserstein, lu2019dcgan}. 

One of the most often used domain adaptation methods is representation adaptation which to align the distribution of the representations from the source domain and target domain by reducing the distribution discrepancy. Some neural networks are build for this, such as feature-based transfer neural network (FTNN) \cite{yang2019intelligent}, deep convolutional transfer learning network (DCTLN) \cite{guo2018deep}. Shao et al. proposed a CNN-based machine fault diagnosis framework in parameter transfer \cite{shao2018highly}, and experimental results show that DCTLN can get the average accuracy of $86.3\%$. Experimental results illustrate that the proposed method can achieve the test accuracy near $100\%$ on three mechanical datasets, and in the gearbox dataset, the accuracy can reach $99.64\%$. 

In adversarial-based domain adaptation, Cheng et al. proposed Wasserstein distance based deep transfer learning (WD-DTL) \cite{cheng2020wasserstein} which uses CNN as pre-trained model. Experimental results show that the transfer accuracy of WD-DTL can reach $95.75 \%$ on average. Lu et al. develop a domain adaptation combined with deep convolutional generative adversarial network (DADCGAN)-based methodology for diagnosing DC arc faults \cite{lu2019dcgan}. DADCGAN is a robust and reliable fault diagnosis scheme based on a lightweight CNN-based classifier can be achieved for the target domain.

In this paper, we choose the DANN architecture to develop a framework of digital twin-supported fault diagnosis. The main reason is that its architecture is simple and can efficiently capture the features from the source domain and generalize well on the target domain. Moreover, DANN’s adversarial training mechanism enables the model to learn domain-invariant features, making it particularly effective in reducing the distribution discrepancy between source and target domains. Furthermore, DANN performs well with limited labeled data from the target domain, addressing the common challenge of insufficient fault data in practical applications. Its ability to handle complex and nonlinear relationships in data and make DANN a reliable and scalable solution for fault diagnosis.

\section{DANN Model Architecture}\label{sec:dann}

\begin{figure*}[hbt]  % 使用 figure* 环境
  \centering
  \includegraphics[width=0.7\textwidth]{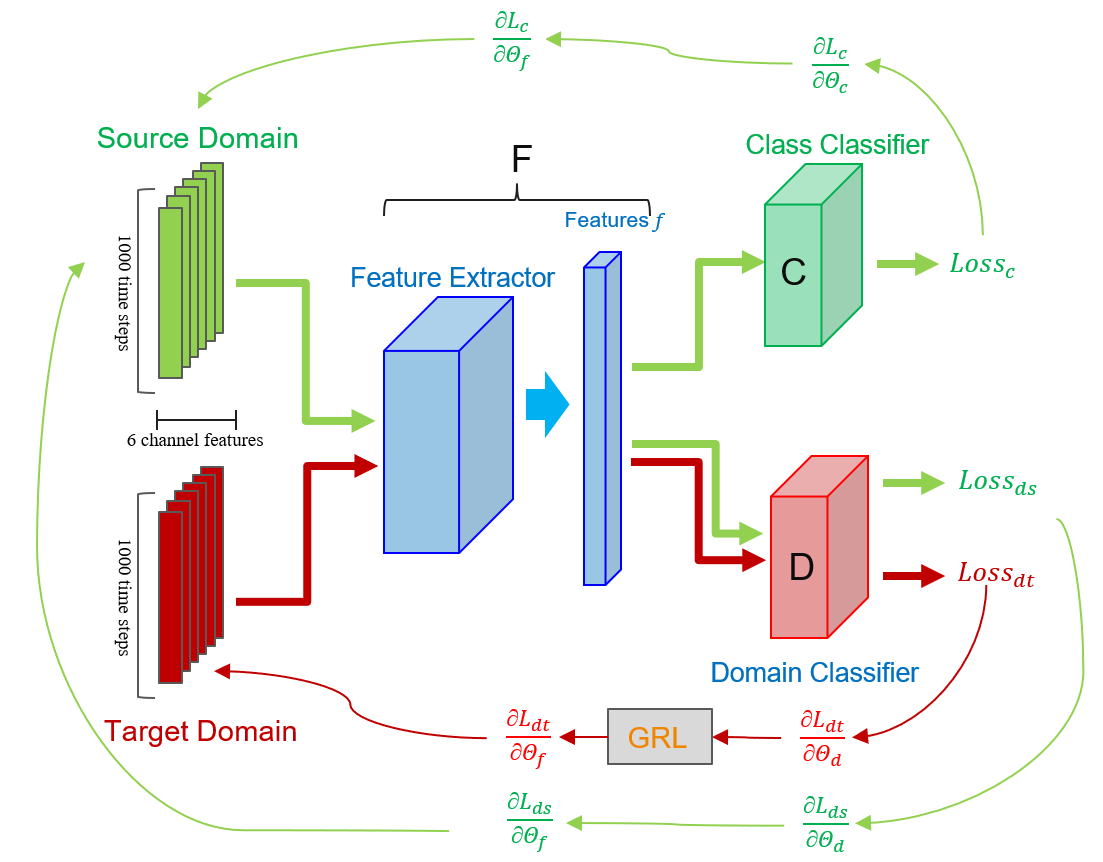} 
 \caption{DANN Architecture \cite{ganin2015unsupervised}}

 \label{fig:DANN}
 \end{figure*}

We use Domain Adversarial Neural Network (DANN) model \cite{ganin2015unsupervised} and extend its application in digital twin in robotics maintenance prediction that previously and originally utilize in transfer learning in domain adaptation. The architecture of DANN is shown in Figure \ref{fig:DANN}. 

Let us assume the input samples are represented by $x \in X$, where $X$ is some input space and certain labels (output) $y$ from the label space $Y$. We assume that there exist two distributions $S(x, y)$ and $T(x, y)$ on $X \otimes Y$ , which will be referred to as the source domain and the target domain. Our goal is to predict labels $y$ given the input $x$ for the target domain. 

We denote with $d_{i}$ the binary variable (domain label) for the $i$th example, which indicates whether $x_{i}$ come from the source domain ($x_{i}$ $\sim$ S(x) if $d_{i}$=0) or from the target distribution ($x_{i}$ $\sim$ T(x) if $d_{i}$=1). We assume that the input $x$ is first representative by a representation learning $G_{f}$ (Feature Extractor) to a d-dimensional feature vector f $\in$ $R^{d}$, and we denote the vector of parameters of all layers in this mapping as $\theta_{f}$, $f$ = $G_{f}$(x; $\theta_{f}$). Then, the feature vector $f$ is representative by $G_{y}$ (label predictor) to the label $y$, and we denote the parameters of this learning with $\theta_{y}$. Finally, the same feature vector $f$ is representative to the domain label $d$ by a mapping $G_{d}$ (domain classifier) with the parameters $\theta_{d}$. 

For the model learning, we minimize the label prediction loss on the annotated part (i.e. the source part) of the train set, also the parameters of both the feature extractor and the label predictor are optimized in order to minimize the empirical loss for the source domain samples. This ensures the discriminativeness of the features $f$ and the overall good prediction performance of the combination of the feature extractor and the label predictor on the source domain. By doing so, we make the features $f$ domain-invariant. 

We need to make the distributions $S(f)$ = {$G_{f}$ (x; $\theta_{f}$)| x$\sim$S(x)} and $T(f)$ = {$G_{f}$ (x; $\theta_{f}$)| x$\sim$T(x)} to be similar \cite{shimodaira2000improving}. To Measure the dissimilarity of the distributions $S(f)$ and $T(f)$, the distributions are constantly changing in learning progresses, we estimate the dissimilarity is to look at the loss of the domain classifier $G_{d}$, provided that the parameters $\theta_{d}$ of the domain classifier have been trained to discriminate between the two feature distributions. In training to obtain domain-invariant features, we seek the parameters $\theta_{f}$ of the feature representative that maximize the loss of
the domain classifier (by making the two feature distributions as similar as possible), and simultaneously seeking the parameters $\theta_{d}$ of the domain classifier that minimize the loss of the domain classifier. And we seek to minimize the loss of the label predictor. The function is:
% \begin{small}
\begin{equation}
\begin{split}
E(\theta_{f}, \theta_{y}, \theta_{d}) = \sum_{i=1..N} L_{y}(G_{y}(G_{f}(x_{i}; \theta_{f}); \theta_{y}), y_{i}) - \\ 
\lambda \sum_{i=1..N} L_{d}(G_{d}(G_{f}(x_{i}; \theta_{f}); \theta_{d}), y_{i}) \\
= \sum_{i=1..N} L_{y}^{i}(\theta_{f}, \theta_{y}) - \lambda \sum_{i=1..N} L_{d}^{i}(\theta_{f}, \theta_{d}) 
\end{split}
\end{equation}
% \end{small}
where $L_{y}$ is the loss for label prediction, $L_{d}$ is the loss for the domain classification, and $L_{y}^{i}$, $L_{d}^{i}$ denote the corresponding loss functions evaluated at the i training example.

We seek the the parameters ${\hat\theta_{f}}$, ${\hat\theta_{y}}$, ${\hat\theta_{d}}$ by solving the following optimization problem:
\begin{equation}
\begin{split}
(\hat\theta_{f}, \hat\theta_{y}) = arg{\min_{\theta_{f}, \theta_{y}}} E(\theta_{f}, \theta_{y}, \hat\theta_{d}) \\
\hat\theta_{d} = arg{\max_{\theta_{d}} E(\hat\theta_{f}, \hat\theta_{y}, \theta_{d})}
\ 
\end{split}
\end{equation}

Then, we do optimization in backpropagation to seek the parameters ${\hat\theta_{f}}$, ${\hat\theta_{y}}$, ${\hat\theta_{d}}$ at the end progressing of class classifier and domain classifier. We also do a gradient reversal layer (GRL) to update and diferences the   
$-\lambda$ facotor in (1). The backpropagation processing passes through the GRL, by the partial derivatives of the loss that is downstream the GRL (i.e. $L_{d}$) w.r.t. the layer parameters that are upstream the GRL (i.e. $\theta_{f}$) get multiplied by $-\lambda$ (i.e. $\frac{\partial L_{d}}{\partial \theta_{f}}$ is effectively replaced with $-\lambda$$\frac{\partial L_{d}}{\partial \theta_{f}}$). We have forward and backward function $R_{\lambda}(x)$:
\begin{equation}
R_{\lambda}(x) = x
\end{equation}

\begin{equation}
\frac{d R_{\lambda}}{dx} = -\lambda I
\end{equation}
where I is the identity matrix.

In feature extractor, we use CNN to do feature extracting, based on our baseline, CNN model has a better results, so we use CNN architecture and its representation to do feature extracting. The CNN is in two convolutional layers, and we set kernel size is 3, number of filters is 64.

\section{Experiments}

\subsection{Dataset}

In this case study, we work on the dataset originally reported in \cite{court2024use}. As \cite{court2024use}, we retained the desired and realized trajectory coordinates (x, y, z) and introduced a derived feature set representing the residuals between the desired and realized trajectories. As a result, the final feature set comprises six features: the desired trajectory coordinates (x, y, z) and the corresponding residuals (x, y, z). 

The source domain dataset generated by the digital twin consists of 3600 samples across 9 distinct labels, with each label containing 400 samples. The real-world measurements are treated as target domain. We have 90 samples in the target domain. We split the source domain dataset into training and validation sets with a $9$ to $1$ ratio, and the target domain dataset is used as the test set.

The DANN described in Sect. \ref{sec:dann} is used to train a fault diagnosis model using the source domain data. Only the measured features in the target domain, but not the labels are used in the training process of the DANN to learn the domain invariate features. Then, the trained DANN is applied to predict the failure labels of the target domain.

\subsection{Evaluation Metrics}
The performance of all methods is evaluated using \textbf{Accuracy} and \textbf{F1 Score}, which are defined as follows:

\paragraph{Accuracy}
\begin{equation}
\begin{aligned}
    \text{Accuracy} &= \frac{\text{Number of Correct Predictions}}{\text{Total Number of Predictions}} \\
    &= \frac{TP + TN}{TP + TN + FP + FN}
\end{aligned}
\end{equation}
where \(TP\), \(TN\), \(FP\), and \(FN\) represent the number of true positives, true negatives, false positives, and false negatives, respectively.

\paragraph{F1 Score}
The F1 Score is the harmonic mean of precision and recall:
\begin{equation}
    \text{F1 Score} = 2 \cdot \frac{\text{Precision} \cdot \text{Recall}}{\text{Precision} + \text{Recall}}
\end{equation}
where:
\begin{equation}
    \text{Precision} = \frac{TP}{TP + FP}, \quad
    \text{Recall} = \frac{TP}{TP + FN}
\end{equation}
These metrics provide a balanced evaluation of the model's performance.

\subsection{Benchmarked models}
We use four current prevalent deep learning methods and models as baseline:
\begin{itemize} 
\item LSTM~\cite{hochreiter1997long} Long Short-Term Memory (LSTM) deals with time series data in deep learning, it often uses for preventing gradient vanishing and gradient explosion in deep learning. ‌LSTM is a special type of recurrent neural network (RNN), and can effectively capture and process long-term dependencies in sequence data by introducing memory units and gating mechanisms.  

\item  Transformer~\cite{vaswani2017attention} Transformer is better at context dependency. And it is very versatile especially in multimodal. This ability to dynamically focus on relevant parts of the input is a key reason why Transformer model excel when processing sequence data.

\item CNN~\cite{lecun1989handwritten} Convolutional Neural Networks (CNN) is mainly used as a visual neural network, which mainly extracts features layer by layer through multiple and deep convolution.

\item TCN~\cite{bai2018empirical} It is a deep learning model specifically designed to process sequential data, combining the parallel processing capabilities of convolutional neural networks (CNN) with the long-term dependent modeling capabilities of recurrent neural networks (RNN).
\end{itemize}

\subsection{Implementation Details}
The implementation of the DANN is carried out using PyTorch. The experiments are conducted on NVIDIA RTX 3060 GPU with the following parameter settings:
Learning rate is 0.001, Batch size is 32, Number of epochs is 250, Optimizer is Adam, and Alpha:

\begin{equation}
    \alpha = \frac{2}{1 + e^{-10 p}} - 1
\end{equation}
where
\begin{equation}
  p = \frac{\text{epoch}}{\text{max epoch}}
\end{equation}

\section{Results and discussions}

\subsection{Average accuracy and F1 score over all methods}

In this subsection, we systematically compare the results from the DANN with the four benchmarked models. We conduct experiments to evaluate the accuracy of the models on the train set, validation set, and real test set, as shown in table \ref{table:accuracy_results}. Additionally, we record the F1-score for each one of the nine classes, as shown in table \ref{table:f1_results}. Due to the randomness of deep learning models, each experiment is conducted five times, and both the average values and standard deviations of the performance metrics are calculated. 

From Table \ref{table:accuracy_results}, it can be seen that the four benchmarked deep learning models do not perform well, especially on the test set. The performance on the test set drops significantly compared to the training set and validation set. This can be explained by the imprecision of the simulation model used to generate the training data. The DANN, on the other hand, achieve much better performance on the test set. This is because through domain adaptation, the DANN is able to extract domain invariate features and generalize them to the target domain.

It is observed from Table \ref{table:f1_results} that most of the benchmarked models exhibit very low classification accuracy for the state healthy. This is because, healthy state is very similar to other states where one motor has steady-state errors. When the simulation model is not accurate, the generated training data are even more difficult to distinguish between healthy and steady-state error states. The DANN, on the other hand, performs well in classifying the state of healthy. This is because after the domain adaptation, in the extracted feature space, the healthy state becomes well-seperated with the other states.

In summary, among the commonly used deep learning models in our experiments, the model that combines a deeper and wider CNN as the backbone with the DANN structure is the relatively optimal choice.

\begin{table*}[htbp]
  \centering
  \caption{Performance Comparison of Baseline Models}
  \label{table:accuracy_results}
  \begin{tabular}{lccc}
    \hline
    Model & Training Accuracy (\%) & Validation Accuracy (\%) & Test Accuracy (\%) \\
    \hline
    LSTM & 96.06±5.57  & 92.22±4.60 & 56.00±4.59 \\
    Transformer & 97.73±0.33 & 75.94±1.52 & 48.44±2.29 \\
    TCN & 87.96±0.86 & 67.67±0.65 & 44.22±1.63 \\
    CNN & \textbf{99.94±0.11} & \textbf{96.78±0.76} & 70.00±1.99 \\
    \textbf{DANN} & 99.29±0.67 & 95.28±0.72 & \textbf{80.22±1.78} \\
    \hline
  \end{tabular}
\end{table*}

\begin{table*}[htbp]
  \centering
  \caption{Performance Comparison on Each Category (F1 Score)}
  \label{table:f1_results}
  \begin{tabular}{lcccccc}
    \hline
    & LSTM & Transformer & TCN & CNN & \textbf{DANN} \\
    \hline 
    Healthy & 0.00±0.00 & 0.00±0.00 & 0.07±0.09 & 0.07±0.09 & \textbf{0.67±0.04} \\
    Motor 1 Stuck & \textbf{0.86±0.06} & 0.63±0.05 & 0.65±0.04 & 0.81±0.03 & 0.84±0.04 \\
    Motor 1 Steady state error & 0.55±0.14 & 0.67±0.09 & 0.46±0.04 & 0.85±0.03 & \textbf{0.90±0.05} \\
    Motor 2 Stuck & 0.72±0.05 & 0.65±0.14 & 0.36±0.03 & 0.73±0.07 & \textbf{0.79±0.04} \\
    Motor 2 Steady state error & 0.53±0.16 & 0.40±0.05 & 0.46±0.08 & \textbf{0.90±0.05} & 0.87±0.02 \\
    Motor 3 Stuck & 0.55±0.05 & 0.54±0.08 & 0.48±0.05 & 0.63±0.09 & \textbf{0.80±0.03} \\
    Motor 3 Steady state error & 0.63±0.11 & 0.38±0.10 & 0.62±0.10 & \textbf{0.91±0.03} & 0.91±0.06 \\
    Motor 4 Stuck & 0.49±0.06 & 0.42±0.08 & 0.40±0.07 & 0.59±0.06 & \textbf{0.78±0.04} \\
    Motor 4 Steady state error & 0.43±0.05 & 0.41±0.07 & 0.28±0.02 & 0.53±0.02 & \textbf{0.62±0.08} \\
    \hline
  \end{tabular}
\end{table*}

\subsection{Ablation study for Digital Twin}

To demonstrate the necessity of using a digital twin model for this task, we conduct an ablation experiment. We train the model using only the real test set, excluding the train and validation sets generated entirely by the digital twin model. In the real test data, we split the dataset into train and testing sets at a ratio of 7:3. Our dataset contains only 90 real data points, and it is clear that most deep learning models struggle to fit on such a small dataset. The results we recorded in Table \ref{table:Ablation}, which indicate that, with such a limited amount of data, common methods cannot make accurate predictions. Use digital twin model to generate simulation data, on the other hand, clearly improve the performance, as the generated simulation data help the deep learning model to better learn the relevant features.

\begin{table*}[htbp]
  \centering
  \caption{Performance Ablation Study}
  \label{table:Ablation}
  \begin{tabular}{lccc}
    \hline
    Model & Only Real Data Accuracy (\%) & Digital twin-supported deep learning (\%) \\
    \hline
    LSTM & 14.92±4.09   & 56.00±4.59 \\
    Transformer & \textbf{18.10±2.58}  & 48.44±2.29 \\
    TCN & 15.24±1.62  & 44.22±1.63 \\
    CNN & 13.97±2.54  & 70.00±1.99 \\
    \textbf{DANN} & 15.87±4.71  & \textbf{80.22±1.78} \\
    \hline
  \end{tabular}
\end{table*}

\section{Conclusions and Future Works}

In this paper, we proposed a new deep learning baseline for fault diagnosis using an existing digital twin dataset. We applied commonly used lightweight deep learning models and demonstrated that the Domain-Adversarial Neural Network (DANN) approach with a CNN backbone, as a transfer learning method, achieves higher accuracy compared to other models. Furthermore, our experiments validate that combining digital twin simulation with domain adaptation techniques can effectively address the issue of limited real-world data in fault diagnosis tasks.

We selected lightweight models such as CNN, TCN, Transformer, and LSTM due to their wide adoption in time-series fault diagnosis, ease of training, and relatively low computational cost. Although these models serve as strong baselines, we acknowledge that more advanced architectures—such as pre-trained large-scale models or graph-based neural networks—may offer improved generalization and performance. Exploring these alternatives remains a promising direction for future research.

However, several limitations remain. First, the DANN framework requires more computational resources and deep learning expertise, which may pose challenges for practical deployment, particularly in resource-constrained industrial settings. Second, the inevitable discrepancies between the digital twin and the real-world system limit the performance of the model, as current simulations cannot fully capture complex physical dynamics. Third, while DANN improves generalization, the deep learning models used in this study still have room for improvement. Future work could explore more robust and generalizable models, such as those pre-trained on large-scale datasets or more advanced domain adaptation methods.

\section*{Acknowledgment}

The research of Zhiguo Zeng is supported by ANR-22-CE10-0004, and chair of Risk and Resilience of Complex Systems (Chaire EDF, Orange and SNCF). Haiwei Fu and Zhenling Chen participate in this project as lab project in their master curricum in Centralesupélec. The authors would like to thank Dr. Myriam Tami for managing this project.

{\scriptsize
\bibliography{ref}
\bibliographystyle{ieeetr}
}

\nolinenumbers

\end{document}